%% file: arxiv.tex
\documentclass[10pt,twocolumn,letterpaper]{article}

\usepackage{cvpr}              

\input{preamble}

\definecolor{cvprblue}{rgb}{0.21,0.49,0.74}
\usepackage[pagebackref,breaklinks,colorlinks,allcolors=cvprblue]{hyperref}

\def\paperID{5205} 
\def\confName{CVPR}
\def\confYear{2025}

\title{EdgeTAM: On-Device Track Anything Model}

\author{\textbf{Chong Zhou$^{1,2}$\thanks{Work done during the internship at Meta Reality Labs.}, Chenchen Zhu$^{1}$, Yunyang Xiong$^{1}$, Saksham Suri$^{1}$, Fanyi Xiao$^{1}$, Lemeng Wu$^{1}$}\\
\textbf{Raghuraman Krishnamoorthi$^{1}$, Bo Dai$^{3}$, Chen Change Loy$^{2}$, Vikas Chandra$^{1}$, Bilge Soran$^{1}$}\\
$^{1}$Meta Reality Labs,
$^{2}$Nanyang Technological University, 
$^{3}$Shanghai AI Laboratory \\
}

\begin{document}
\maketitle
\input{sec/0_abstract}    
\input{sec/1_intro}
\input{sec/2_related}
\input{sec/3_method}
\input{sec/4_expriement}
\input{sec/5_conclusion}
\input{sec/X_suppl}

{
    \small
    \bibliographystyle{ieeenat_fullname}
    \bibliography{main}
}


\end{document}

%% file: preamble.tex
\usepackage{comment}

\newcommand{\f}[1]{\textbf{#1}} 

\newcommand{\inlinesection}[1]{\noindent \textbf{#1}$\,\,\,$}

\usepackage{pifont}
\newcommand{\cmark}{\ding{51}}

\usepackage{multirow}
\usepackage{makecell}
\usepackage[table]{xcolor}

\newcommand{\jnf}{$\mathcal{J}$\&$\mathcal{F}$}
\newcommand{\g}{$\mathcal{G}$}

\newcommand{\crs}{\textsuperscript{\textdagger}}

\usepackage{float}

%% file: sec/0_abstract.tex
\begin{abstract}
On top of Segment Anything Model (SAM), SAM 2 further extends its capability from image to video inputs through a memory bank mechanism and obtains a remarkable performance compared with previous methods, making it a foundation model for video segmentation task. In this paper, we aim at making SAM 2 much more efficient so that it even runs on mobile devices while maintaining a comparable performance. Despite several works optimizing SAM for better efficiency, we find they are not sufficient for SAM 2 because they all focus on compressing the image encoder, while our benchmark shows that the newly introduced memory attention blocks are also the latency bottleneck. Given this observation, we propose EdgeTAM, which leverages a novel 2D Spatial Perceiver to reduce the computational cost. In particular, the proposed 2D Spatial Perceiver encodes the densely stored frame-level memories with a lightweight Transformer that contains a fixed set of learnable queries. Given that video segmentation is a dense prediction task, we find preserving the spatial structure of the memories is essential so that the queries are split into global-level and patch-level groups. We also propose a distillation pipeline that further improves the performance without inference overhead. As a result, EdgeTAM achieves 87.7, 70.0, 72.3, and 71.7 \jnf{} on DAVIS 2017, MOSE, SA-V val, and SA-V test, while running at 16 FPS on iPhone 15 Pro Max.

\end{abstract}

%% file: sec/1_intro.tex
\section{Introduction}
\label{sec:intro}

\input{figure/0_fig_teaser}

Segment Anything Model (SAM) \cite{sam} is the first foundation model for promptable image segmentation. Various studies show its magnificent capabilities on zero-shot generalization and transfer learning \cite{sam-eval1,sam-eval2,sam-eval3,sam-eval4}. On top of SAM, recently, SAM 2 \cite{sam2} extends the original SAM to handle both image and video inputs, with a memory bank mechanism, and is trained with a new large-scale multi-grained video tracking dataset (SA-V).


\input{figure/1_fig_benchmark}

Despite achieving an astonishing performance compared to previous video object segmentation (VOS) models and allowing more diverse user prompts, SAM 2, as a server-side foundation model, is not efficient for on-device inference. For instance, the smallest SAM 2 variant runs at only around 1 FPS on an iPhone 15 Pro Max \footnote{We convert to CoreML model with coremltools \cite{coremltools} and benchmark with CPU and NPU. Throughout the paper, we interchangeably use iPhone and iPhone 15 Pro Max for simplicity.}.
Furthermore, existing methods \cite{mobilesam,efficientsam,edgesam} that optimize SAM for better efficiency only consider squeezing its image encoder since the mask decoder is extremely lightweight. But as shown in Fig.~\ref{fig:benchmark}, this is not sufficient for SAM 2 because even when the image encoder is replaced with much more compact visual backbones, such as ViT-Tiny \cite{deit} and RepViT \cite{repvit}, the latency does not improve by much due to the computationally demanding memory attention blocks that are newly introduced in SAM 2.
Specifically, SAM 2 encodes past frames with a memory encoder, and these frame-level memories together with object-level pointers (obtained from the mask decoder) serve as the memory bank. These are then fused with the features of current frame via memory attention blocks. As these memories are densely encoded, this leads to a huge matrix multiplication during the cross-attention between current frame features and memory features. Therefore, despite containing relatively fewer parameters than the image encoder, the computational complexity of the memory attention is not affordable for on-device inference. The hypothesis is further proved by Fig.~\ref{fig:benchmark}, where reducing the number of memory attention blocks almost linearly cuts down the overall decoding latency and within each memory attention block, removing the cross attention gives the most significant speed-up.

To make such a video-based tracking model run on device, in EdgeTAM, we look at exploiting the redundancy in videos. To do this in practice, we propose to compress the raw frame-level memories before performing memory attention. We start with na\"{i}ve spatial pooling and observe a significant performance degradation, especially when using low-capacity backbones. To mitigate this issue, we turn to learning-based compressors such as Perceiver~\cite{perceiver,perceiverio}, which summarizes the dense feature map with a small fixed set of learned queries. However, na\"{i}vely incorporating a Perceiver also leads to a severe drop in performance. We hypothesize that as a dense prediction task, the video segmentation requires preserving the spatial structure of the memory bank, which a na\"{i}ve Perceiver discards.

Given these observations, we propose a novel lightweight module that compresses frame-level memory feature maps while preserving the 2D spatial structure, named 2D Spatial Perceiver. Specifically, we split the learnable queries into two groups, where one group functions similarly to the original Perceiver, where each query performs global attention on the input features and outputs a single vector as the frame-level summarization. In the other group, the queries have 2D priors, \emph{i.e.}, each query is only responsible for compressing a non-overlapping local patch, thus the output maintains the spatial structure while reducing the total number of tokens.
As a plug-in module, 2D Spatial Perceiver can be integrated with any variants of SAM 2 and speed up the memory attention by $8\times$ with comparable performance. For instance, when using RepViT-M1 \cite{repvit} as the backbone and two memory attention blocks, leveraging the 2D Spatial Perceiver yields 16 FPS on iPhone, which is $6.4 \times$ faster than the baseline and even surpasses it on the challenging SA-V val set \cite{sam2} by 0.9 \jnf{}.

In addition to the architecture improvement, we further propose a distillation pipeline that transfers the knowledge of the powerful teacher SAM 2 to our student model, which improves the accuracy at no cost of inference overhead.
Specifically, the training procedure of SAM 2 has two stages, where firstly the model is trained with the promptable image segmentation task on SA-1B~\cite{sam} with memory-related module detached, then in the second stage, it is trained with all modules included for the promptable video segmentation task on both SA-1B and SA-V \cite{sam2} datasets.
We find that in both stages, aligning the features from image encoders of the original SAM 2 and our efficient variant 
benefits the performance. Besides, we further align the feature output from the memory attention between the teacher SAM 2 and our student model in the second stage so that in addition to the image encoder, memory-related modules can also receive supervision signals from the SAM 2 teacher.
As a result, with the proposed distillation pipeline,  we improve the \jnf{} on SA-V val and test by 1.3 and 3.3, respectively.

Putting together, we propose \textbf{EdgeTAM} (Track Anything Model for Edge devices), that adopts a 2D Spatial Perceiver for efficiency and knowledge distillation for accuracy.
Our contributions can be summarized in the following:
\begin{itemize}
    \item Through comprehensive benchmark, we reveal that the latency bottleneck lies in the memory attention module.
    \item Given the latency analysis, we propose a 2D Spatial Perceiver that significantly cuts down the memory attention computational cost with comparable performance, which can be integrated with any SAM 2 variants.
    \item We experiment with a distillation pipeline that performs feature-wise alignment with the original SAM 2 in both the image and video segmentation stages and observe performance improvements without any additional cost during inference.
    \item The resulting EdgeTAM can run at \textbf{16 FPS} on an iPhone, which is notably faster than existing video object segmentation models and surpasses or is on par with the previous state-of-the-art methods. To our knowledge, it is the first model running on device for the task of unified segmentation and tracking.
\end{itemize}

%% file: figure/0_fig_teaser.tex
\begin{figure}[t]
    \centering
    \includegraphics[width=0.5\textwidth]{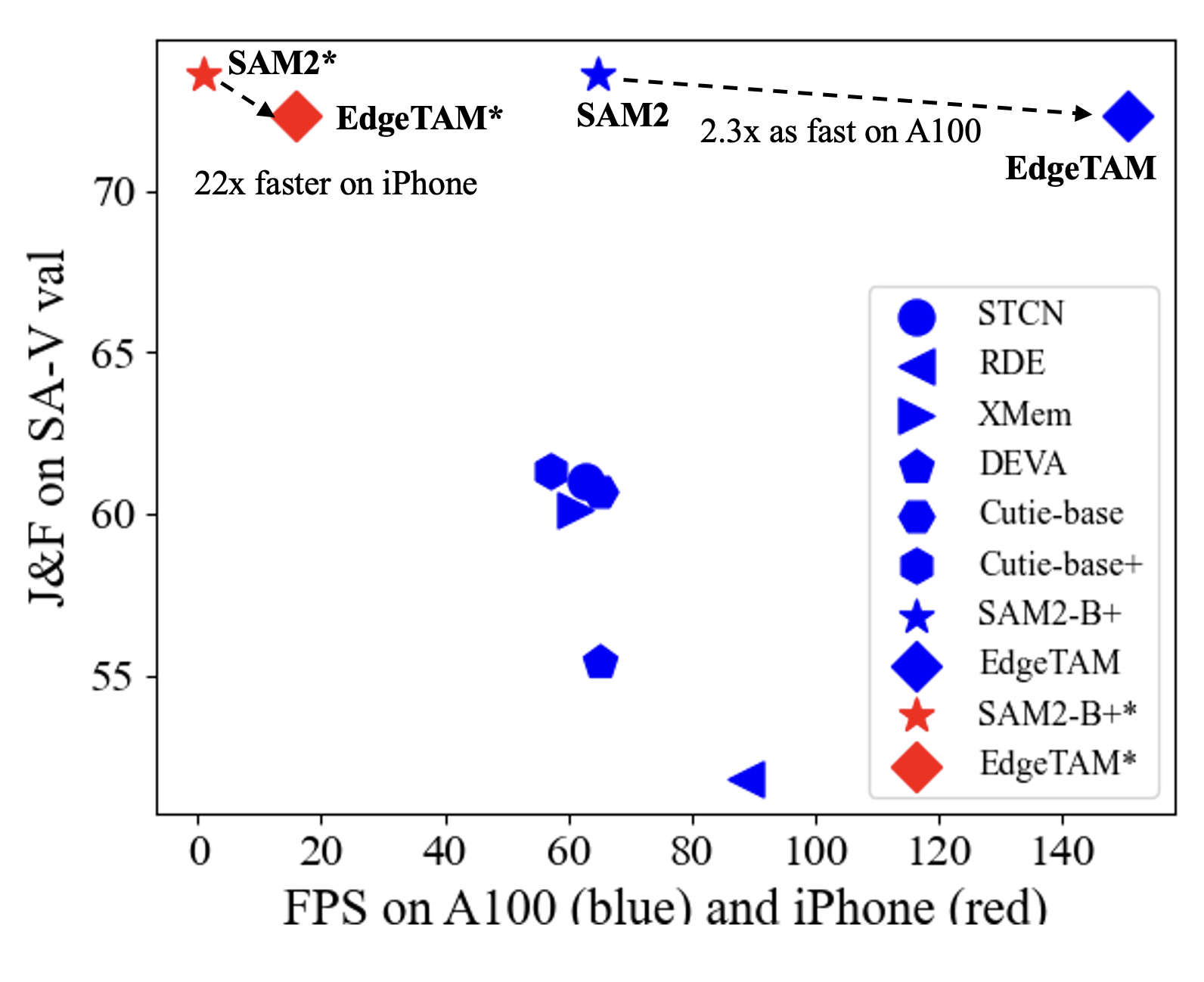}
    \caption{\textbf{Speed-performance trade-offs on iPhone 15 Pro Max and NVIDIA A100.} EdgeTAM is significantly faster than SAM 2 on edge devices and compare to other VOS methods, it is also more accurate on the challenging SA-V val dataset. Note that, EdgeTAM can run at 16 FPS on iPhone 15 Pro Max.}
    \label{fig:teaser}
\end{figure}

%% file: figure/1_fig_benchmark.tex
\begin{figure*}[t]
    \centering
    \includegraphics[width=1.0\textwidth]{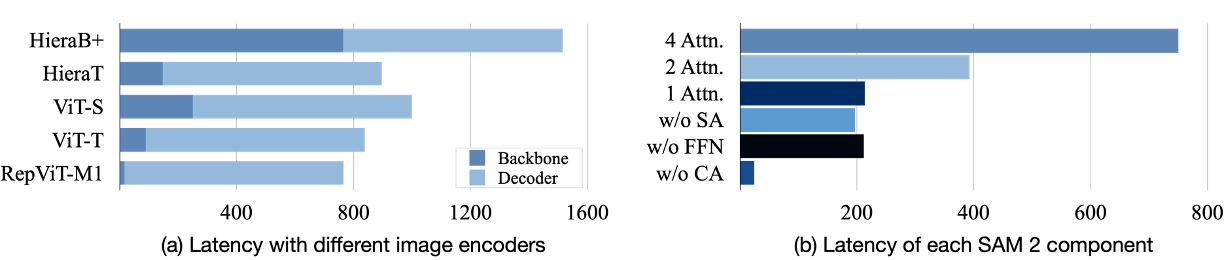}
    \caption{\textbf{Single frame latency (ms) on iPhone.} In (a), we show that only replacing image encoder with more compact backbones is not enough for further speed-up since decoder is also a bottleneck. In (b), through reducing the number of memory attention blocks and removing certain modules, we find that the cross attention (CA) is the root cause.}
    \label{fig:benchmark}
\end{figure*}

%% file: sec/2_related.tex
\section{Related Work}
\label{sec:related}

\inlinesection{Video Object Segmentation (VOS).} The objective of the VOS task is, given the ground-truth (GT) object segmentation mask on the first frame, tracking and predicting the object mask throughout the following frames in the video. Online learning approaches \cite{vos1,vos2,vos3,vos4,vos5,vos6,vos7,vos8,vos9,vos10,vos11,vos12} formulate the task as a semi-supervised learning problem, where during test time, the model is fine-tuned with the GT mask on the first frame. However, this line of work usually suffers from inference inefficiency, being input sensitive and hard to scale up with large amounts of training data. To avoid test-time training, offline-trained models propose to leverage template matching \cite{vos21,vos22,vos23,vos24,vos25,vos26,vos27,vos28}, or memory bank \cite{vos-bank-1,vos-bank-2} to keep track of the identity information in the annotated and predicted frames. In terms of the network architecture, some works adopt recurrent networks for spatial-temporal encoding \cite{vos-rnn,vos-rnn-2,vos-rnn-3,rde}, while recently, Transformer-based models \cite{rde,stcn,aot,deaot,isvos,jointformer,simvos,vos31,vos32,vos33,xmem,xmem++,cuite} demonstrate better performance.

\inlinesection{Segment Anything Model (SAM).} SAM \cite{sam} defines a new prompt-based segmentation task where the user prompts can be points, boxes, and masks. SAM 2 \cite{sam2} further extends the task to the video inputs, namely promptable video segmentation (PVS). Different from VOS, users can provide annotations at any frame and at multiple time steps with any combination of SAM prompts, making VOS a special case of PVS. Both SAM and SAM 2 follow the same meta architecture of image encoder and prompt-based mask decoder, but to capture temporal information, SAM 2 supplements a memory banking mechanism. Thanks to training on diverse and large-scale datasets, SA-1B \cite{sam} and SA-V \cite{sam2}, SAM excels in both general perception and downstream tasks \cite{sam-eval1,sam-eval2,sam-eval3,sam-eval4,sam2-eval1,sam2-eval2}. To make SAM more efficient and more friendly to low-capacity devices, several works \cite{mobilesam, efficientsam, edgesam, fastsam, repvitsam} propose to squeeze its image encoder to more compact visual backbones with knowledge distillation and/or masked image pre-training. However, through our benchmark, we find that apart from the image encoder, the newly introduced memory-related modules in SAM 2 are also the speed bottleneck; thus, replacing the image encoder is no longer sufficient. Therefore, we propose a novel plug-in module to accelerate memory fusion to address the problem, together with a distillation pipeline adapted for video inputs.

%% file: sec/3_method.tex
\section{Methodology}
\label{sec:method}

In this section, we first briefly introduce the Segment Anything Model 2 (SAM 2), which our model is based on. Then, we propose our architecture-level improvements and knowledge distillation pipeline, respectively.

\subsection{Preliminary: SAM 2}
Overall, SAM 2 consists of four components, namely image encoder $E_\text{img}$, mask decoder $D$, memory encoder $E_\text{mem}$, and memory attention $A$, with the former two almost identical to the original SAM except for the skip connection between the two. In particular, $E_\text{img}$ is a hierarchical backbone called Hiera \cite{hiera}, which outputs feature maps in three different strides, 4, 8, and 16 denoted by $F_4$, $F_8$, $F_{16}$, respectively:
\begin{equation}
\{F_4, F_8, F_{16} \} = E_\text{img}(I),
\label{eq:1}
\end{equation}
where $I$ is the current frame input. Then, $F_{16}$ is fused with memory features $\{M_1, M_2, \dots, M_T \}$ \footnote{For simplicity, $M$ denotes the frame-level memory feature map and we omit the object pointers (vectors from the mask decoder), which add negligible computational cost.} from previous $T$ frames with the memory attention $A$. The memory attention is essentially a stack of Transformer \cite{transformer} blocks. In this setup, $F_{16}$ serves as the queries, while memory features, concatenated along the temporal dimension, provide the keys and values:
\begin{equation}
F_M = A(F_{16}, M_1, M_2, \dots, M_T),
\label{eq:2}
\end{equation}
where $F_M$ is the image feature conditioned on memories. Next, mask decoder $D$ encodes the user prompt and decodes the mask prediction $O$ given the prompt embedding $P$ and image features $F_M$, $F_4$, $F_8$:
\begin{equation}
O = D(F_M, F_4, F_8, P).
\label{eq:3}
\end{equation}
Finally, $F_{16}$ and $O$ are fused and encoded with the memory encoder $E_\text{mem}$ and enqueued the memory bank in a first-in-first-out manner:
\begin{equation}
M_{T+1} = E_\text{mem}(F_{16}, O).
\label{eq:4}
\end{equation}

\input{figure/2_fig_arch}

\subsection{EdgeTAM}
\inlinesection{Na\"ive Adaptations.} As shown in Fig.~\ref{fig:arch}, the meta architecture of SAM 2 follows closely with SAM, whose image encoder is the heaviest component in terms of parameters and computation. While the newly introduced memory-related module takes up only a small proportion of the total parameters, our benchmark (Fig. \ref{fig:benchmark}) shows that memory attention is also a latency bottleneck. Therefore, a na\"ive technique to push for improved efficiency is to substitute the image encoder with compact backbones and to reduce the number of memory attention blocks. To this end, following EdgeSAM \cite{edgesam}, we opt for RepViT-M1~\cite{repvit} as the backbone and decrease the memory attention from 4 to 2 blocks. However, the inference throughput is still far from being satisfactory when deployed on mobile devices (merely 2.5 FPS on iPhone 15 Pro Max). 

Taking a closer look, we observe that each memory feature $M_t$ has the same size as the image feature $F_M \in \mathcal{R}^{C\times H\times W}$, where $C\text{ = }64$, $H\text{ = }W\text{ = }64$ denote channels, height and width respectively. With $T$ frames in the memory bank, the computational complexity of memory attention becomes $\mathcal{O}(TCH^2W^2)$, which translates to a huge matrix multiplication that mobile devices with limited scale of parallelism perform inefficiently. While $T$ is already relatively small compared to other VOS methods, reducing it will lead to the degradation of temporal consistency and occlusion handling. On the other hand, videos are known to be information redundant. Thus, we propose to summarize the memory spatially before performing memory attention.

\inlinesection{Global Perceiver.} Inspired by Perceiver \cite{perceiver,perceiverio}, we encode each memory feature $M_t$ with a stack of attention modules to compress the densely stored memories $M_t \in \mathcal{R}^{C\times H\times W}$ into a small set of vectors $G_t \in \mathcal{R}^{C\times N_g}$, where $N_g$ is the number of learnable latents and $N_g \ll H \times W$. Specifically, we denote the latents as $Z_g \in \mathcal{R}^{C\times N_g}$ and perform single-head cross attention (CA) between $Z_g$ and $M_t$, followed by self attention (SA) as follows:
\begin{equation}
\begin{aligned}
Z_g' &= \text{CA}(Q(Z_g), K(M_t+p), V(M_t+p)), \\
G_t  &= \text{SA}(Z_g'),
\label{eq:5}
\end{aligned}
\end{equation}
where $Q$, $K$, and $V$ represent the projections for query, key, and value in CA, respectively. $Z_g'$ is the intermediate feature and $p$ denotes the positional embeddings \cite{rope}. Here, each latent can attend globally to the memory feature and summarize it into a single vector. While the Global Perceiver introduces negligible inference cost, it cuts down the complexity of the memory attention to $\mathcal{O}(TCHWN_g)$. However, despite adding positional embeddings to the input of Global Perceiver, the resulting compressed memories contain only implicit positional information as the output does not maintain its spatial structure. Meanwhile, as a dense prediction task, video object segmentation requires more explicit positional information \cite{sam2} and local features \cite{vos33}. We thus further propose a 2D Spatial Perceiver for this purpose.

\input{figure/3_fig_distill}

\inlinesection{2D Spatial Perceiver.} Similar to the Global Perceiver, 2D Spatial Perceiver shares the same network architecture and parameters. However, we assign spatial prior to the learnable latents $Z_l \in \mathcal{R}^{C\times N_l}$ and restrict each latent to only attend to a local window. Specifically, we perform the window partition \cite{swin} to split the memory feature map into $N_l$ non-overlapping patches, and move the positional embedding $p'$ from input to output $L_t$:
\begin{equation}
\begin{aligned}
M_t' &= \text{window\_partition}(M_t) , \\
Z_l' &= \text{CA}(Q(Z_l), K(M_t'), V(M_t')), \\
L_t' &= \text{SA}(Z_l'), \\
L_t  &= \text{window\_unpartition}(L_t)+p'.
\label{eq:6}
\end{aligned}
\end{equation}
The different designs of Global and 2D Spatial Perceiver encourage different behaviors, where global latents $Z_g$ have certain redundancy (multiple latents attend to the same input) and can dynamically distribute all over the image whereas 2D latents $Z_l$ are forced to deal with local patches. And both possess desirable merits for feature summarization. Therefore, we combine them by flattening along the spatial dimension and concatenating along the flattened dimension. Note that, our implementation stacks the blocks in Eq.~\ref{eq:5} and Eq.~\ref{eq:6} twice. Overall, when applying the proposed modules, the complexity of memory attention decreases from $\mathcal{O}(TCH^2W^2)$ to $\mathcal{O}(TCHW(N_g+N_l))$. In practice, we control the speed-up ratio to around $T$ times, \emph{i.e.,} $(HW)/(N_g+N_l) \approx T$, so that the self and cross attention blocks in memory attention have similar complexity.

\subsection{Distillation Pipeline}
As shown in Fig.~\ref{fig:distill}, the training pipeline of SAM 2 can be divided into image segmentation pre-training $S_{img}$ and video segmentation training $S_{vid}$ stages. Previous methods \cite{mobilesam,efficientsam,edgesam} demonstrate that knowledge distillation on $S_{img}$ helps improve performance on images. Here, we extend this idea to the video domain and treat the distillation loss as an auxiliary loss, meaning task-specific losses are also implemented during training.

Particularly, during $S_{img}$, we adopt the same task-specific losses $\mathcal{L}_\text{task}$ as SAM (dice loss \cite{dice} and focal loss \cite{focal_loss} for mask prediction and L1 loss for mask confidence prediction) and meanwhile, align the image encoder feature map ($F_{16}$ in Eq.~\ref{eq:1}) between the teacher and student models with MSE loss $\mathcal{L}_\text{img}$. The pre-training loss $\mathcal{L}_\text{sam}$ can be formulated with:
\begin{equation}
\mathcal{L}_\text{sam} = \mathcal{L}_\text{task}(O, \text{GT}) + \gamma \cdot \mathcal{L}_\text{img}(F_{16}^t, F_{16}^s),
\end{equation}
where $O$ is the mask prediction obtained from Eq.~\ref{eq:1} and Eq.~\ref{eq:3}. Here Eq.~\ref{eq:2} is skipped due to the lack of memory bank and $F_M=I$. Here, GT, $\gamma$, $F_{16}^t$ and $F_{16}^s$ denote the ground-truth labels, loss weight, teacher and student image encoder features respectively.

Finally, in stage $S_{vid}$, the task-specific losses include an additional BCE loss for occlusion prediction. Besides, in order to let student's memory-related modules receive supervision from the teacher, apart from $\mathcal{L}_\text{img}$, we add another MSE loss $\mathcal{L}_\text{mem}$ to align the $F_M^t$ and $F_M^s$ from teacher and student (Eq.~\ref{eq:2}). The resulting total loss becomes:
\begin{equation}
\mathcal{L}_\text{sam2} = \mathcal{L}_\text{task}(O, \text{GT}) + \alpha \cdot \mathcal{L}_\text{img}(F_{16}^t, F_{16}^s) + \beta \cdot \mathcal{L}_\text{mem}(F_M^t, F_M^s),
\end{equation}
with $\alpha$ and $\beta$ serving as the loss weights.

%% file: figure/2_fig_arch.tex
\begin{figure*}[t]
    \centering
    \includegraphics[width=1.0\textwidth]{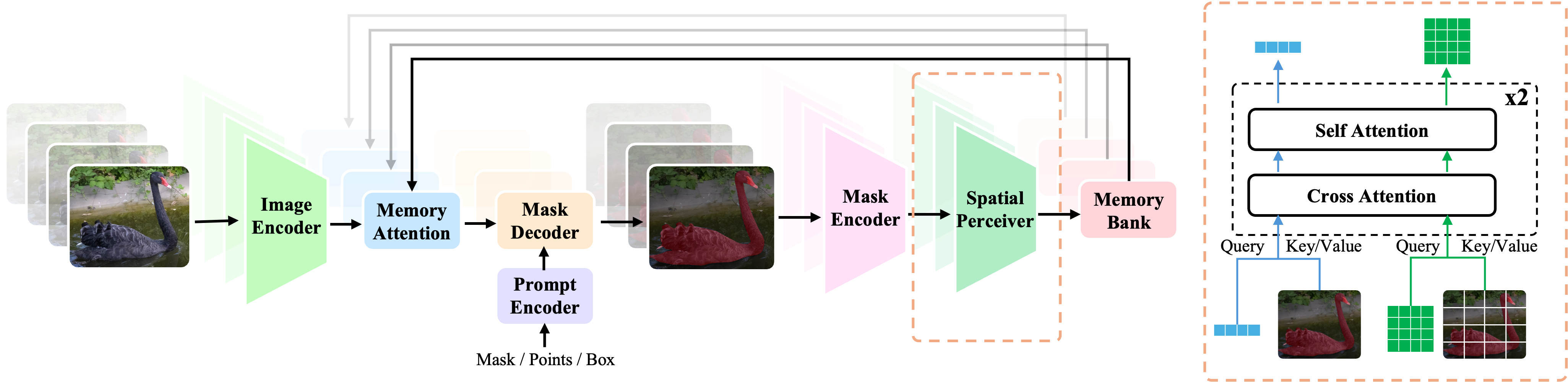}
    \caption{\textbf{Overall architecture of EdgeTAM.} The meta architecture of EdgeTAM follow SAM 2 and the main difference is the proposed plug-in module, 2D Spatial Perceiver, which is marked with orange dotted box.}
    \label{fig:arch}
\end{figure*}

%% file: figure/3_fig_distill.tex
\begin{figure*}[t]
    \centering
    \includegraphics[width=1.0\textwidth]{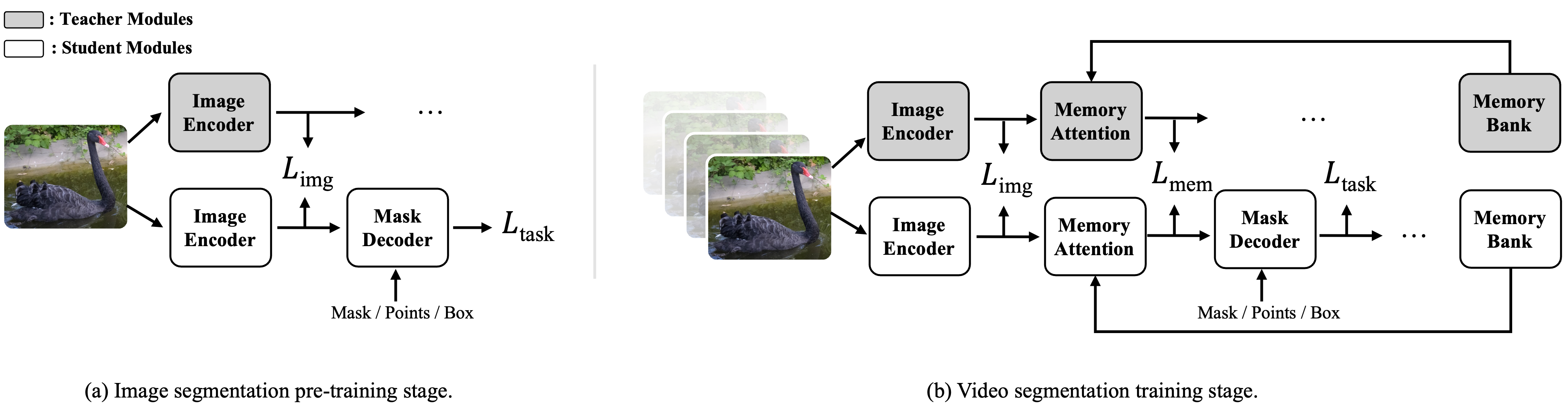}
    \caption{\textbf{The distillation pipeline in EdgeTAM.} In the image pre-training stage, we align the features from teacher's and student's image encoder. And in the video training stage, we additionally align the features output from memory attention between teacher and student. For both stages, task-specific losses are used.}
    \label{fig:distill}
\end{figure*}

%% file: sec/4_expriement.tex
\section{Experiments}
\label{sec:expriment}

\subsection{Implementation Details}
\inlinesection{Training.} In general, the training procedure of EdgeTAM follows SAM 2. We set the input resolution to $1024\times 1024$. During the image segmentation pre-training stage, we train on the SA-1B dataset for 2 epochs with a batch size of 128. We use AdamW \cite{adamw} as the optimizer ($\beta_1,\beta_2\text{=}0.9, 0.999$) and set the learning rate to $4e^{-4}$ with a reciprocal square-root scheduler \cite{sqrt}. We perform $\text{L}_2$ gradient clipping at 0.1 and set weight decay to 0.1. The loss weights for dice, focal, IoU, and $\mathcal{L}_\text{img}$ are 20, 1, 1, and 1, respectively. For each training sample, we allow a maximum of 64 objects and add 7 correction points iteratively. Random horizontal flip is the only data augmentation in this stage. For video segmentation training, we train on SA-V, a 10\% randomly sampled subset of SA-1B, DAVIS, MOSE, and YTVOS for 130K iterations with a 256 batch size. Most configurations follow the previous stage, except that the learning rate equals $6e^{-5}$ for the image encoder and $3e^{-4}$ for others with a cosine scheduler. The loss balancing factor for dice is 20 and 1 for focal, IoU, occlusion, $\mathcal{L}_\text{img}$, and $\mathcal{L}_\text{mem}$. Each video sample contains 8 frames with almost 3 objects and is augmented with horizontal flip, color jitter, affine, and grayscale transformations.

\inlinesection{Progressive fine-tuning with longer training samples.} Following SAM 2.1, we fine-tune the trained EdgeTAM model on 16-frame sequences. During the fine-tuning, we freeze the image encoder and do not apply distillation. The training set is the same as the video segmentation training stage but the total iterations are reduced to 1/3 of the original schedule. Furthermore, given that EdgeTAM consumes much less VRAM than SAM 2, we are able to further fine-tune the 16-frame model with 32-frame training samples with the same schedule. Note that the memory bank size stays the same and only the training samples become longer, so the inference cost remains the same.

\input{figure/4_fig_all_pvs}
\input{table/1_tab_img}

\inlinesection{Model.} By default, we use RepViT-M1 \cite{repvit} pre-trained on ImageNet \cite{imagenet} classification as the image encoder. We also experiment with ViT-Tiny \cite{deit} pre-trained with MAE \cite{mae} on ImageNet. The number of memory attention blocks is 2 and we allocate 256 learnable latents for both Global Perceiver and 2D Spatial Perceiver. The memory bank sizes for frame-level memories and object pointers are 7 and 16 following SAM 2. The positional embeddings of Global Perceiver and 2D Spatial Perceiver are sinusoidal, and 2D-RoPE \cite{rope}, respectively. We use the SAM2-HieraB+ as the teacher with the publicly available checkpoint\footnote{https://github.com/facebookresearch/sam2}.

\subsection{Datasets}
\inlinesection{Training.} We train on SA-1B \cite{sam}, SA-V \cite{sam2}, DAVIS \cite{davis}, MOSE \cite{mose}, and YTVOS \cite{ytvos} datasets. SA-1B contains 11M images with 1.1B mask annotations in diverse granularities (in both part-level and object-level). The average resolution of images in SA-1B is  $3300\times 4950$ pixels. So far, it is the largest dataset available for image segmentation tasks. SA-V follows the criteria of SA-1B and collects 190.9K masklet annotations across 50.9K videos, which have an average duration of 14 seconds with 54\%/46\% indoor/outdoor scenes and are resampled to 24 FPS. Note that, the annotation frame rate is 6 FPS. Besides, 293/278 masklets from 155/150 videos are reserved as the SA-V val/test splits, which are manually picked to focus on challenging cases with fast-moving, complex occlusions, and disappearance.

\inlinesection{Evaluation.} Our evaluation can be split into three settings: (1) Promptable Video Segmentation (PVS), where the user can click on any frames in the video to indicate an object of interest; (2) Segment Anything (SA), which is same as PVS but works with images; (3) Semi-supervised Video Object Segmentation (VOS), where ground-truth masks on the first frame are available during inference. For the video task, we report \jnf{} \cite{davis} and \g{} \cite{ytvos} as the metric and for images, we use mIoU.

For PVS, we evaluate with the zero-shot protocol across 9 datasets with both online and offline modes. For SA, we evaluate on SA-23 \cite{sam}, which consists of 23 open-source datasets in both video (each frame is considered as an image) and image domains. Finally, for VOS, we provide performance on the popular DAVIS 2017 \cite{davis}, MOSE \cite{mose}, and YouTubeVOS \cite{ytvos} val sets and the challenging SA-V val/test set \cite{sam2}.


\input{table/2_tab_vos}
\input{table/3_tab_ablation}

\subsection{Promptable Video Segmentation (PVS)}
One of the key features of EdgeTAM is that it follows the same meta architecture of SAM 2, which enables it to perform promptable video segmentation with various user inputs on any frames. As shown in Fig.~\ref{fig:all_pvs}, we follow the same online and offline PVS settings as SAM 2, which simulate user interaction in the real world. The offline mode allows multiple times of playbacks to only add correction points on the frames with large errors, while the online mode only annotates the frames in a single forward pass. 
Compared to SAM + XMem++ and SAM + Cuite, EdgeTAM outperforms both across all settings with considerable margins. Besides, thanks to being trained in an end-to-end manner and distilled with the SAM 2 teacher the gap becomes larger as the number of annotated frames increases. Besides, even compared with the original SAM 2, EdgeTAM achieves comparable results despite being significantly smaller and faster.

\input{figure/5_fig_vis}

\subsection{Segment Anything (SA)}
Both SAM 2 and EdgeTAM can function as image segmentation models with the memory module detached. As shown in Tab.~\ref{tab:img}, EdgeTAM achieves comparable mIoU performance with SAM and SAM 2, especially with more input points. For example, with five input points, on average, EdgeTAM even surpasses SAM-H (81.7 \emph{v.s.} 81.3), which is dedicated to image segmentation. Note that, our EdgeTAM is not trained with the internal datasets that both SAM 2 and SAM 2.1 use. Given its real-time speed, EdgeTAM can be used as a unified on-device segmentation model for both images and videos.

\subsection{Video Object Segmentation (VOS)}
While EdgeTAM is trained only with the SA-V and SA-1B dataset, as shown in Tab.~\ref{tab:vos}, on MOSE, DAVIS, and YTVOS, it is on par or surpasses previous state-of-the-art VOS models that are trained on these datasets. This demonstrates the robustness of EdgeTAM under the zero-shot setting. More importantly, it is impractical to deploy multiple models on device with one for certain types of data. Also, thanks to training on SA-V, EdgeTAM surpasses all its counterparts except for SAM 2 and SAM 2.1 on SA-V val and test. Note that, the masks in SA-V val/test have different granularities, while those of other datasets are at object-level. This shows the flexibility of EdgeTAM. In addition, for speed benchmarking, our main goal is inference on edge devices and we observe even with torch compile, the streaming multiprocessor utilization of EdgeTAM is still relatively low. Through the Torch profile, we find that on high-end GPU, the CPU (CUDA kernel launching) becomes the bottleneck for EdgeTAM. Thus, we encourage focusing on edge device latency, which EdgeTAM is designed for.

\subsection{Ablations}
For all the ablation studies, we train with one-third of the original training schedule (43k steps). As shown in Tab.~\ref{tab:ablation}(a), we first ablate the effectiveness of each proposed component. In the table, we set the baseline as RepViT-M1 with two memory attention blocks and we also compare with simply downsampling the spatial memories instead of using the 2D Perceiver. Experiments show that 2D Spatial Perceiver is both faster and more accurate than the baseline and 4$\times$4 average pooling (0.4 to 2.7 better). Besides, the proposed distillation pipeline further improves the \jnf{} on SA-V val and test by 1.3 and 3.3. 
Then, in Tab.~\ref{tab:ablation}(b), we vary the number of global and 2D latents and find that using both yields the best performance and speed-up. Note that, using 2D latents speed up the baseline by 6.3 $\times$ with better performance.
Tab.~\ref{tab:ablation}(c) shows using 2D Perceiver on different combinations of image encoders and the number of memory attention blocks. And we opt for RepViT-M1 with two memory attentions for the best trade-off. Finally, in Tab.~\ref{tab:ablation}(d), we study whether to use self attention in the 2D Perceiver network. The motivation here is that as each 2D latent attends to a local patch that has no overlap with each other, incorporating self attention blocks will encourage the communication between 2D latents to yield better features. Our results verify this hypothesis.

\input{table/suppl/2_tab_prompt_vos}

\subsection{Qualitative Results}
In Fig.~\ref{fig:vis}, we compare the visualization results of EdgeTAM and SAM 2 on the YouTubeVOS val dataset. We pick two representative examples, one with multiple instances from the same class gathering together, and the other with a fast-moving object with a large distortion. For the first example, EdgeTAM yields similar results as SAM 2 and keeps the identity of each instance throughout the clip. However, in the second example, we observe that EdgeTAM falls into a typical failure case that the tracking granularity might always follow SAM 2. In the example, EdgeTAM does not include the bird feet in the mask predictions given that in previous frames, the feet are not visible. 

%% file: figure/4_fig_all_pvs.tex
\begin{figure*}[t]
    \centering
    \includegraphics[width=0.9\textwidth]{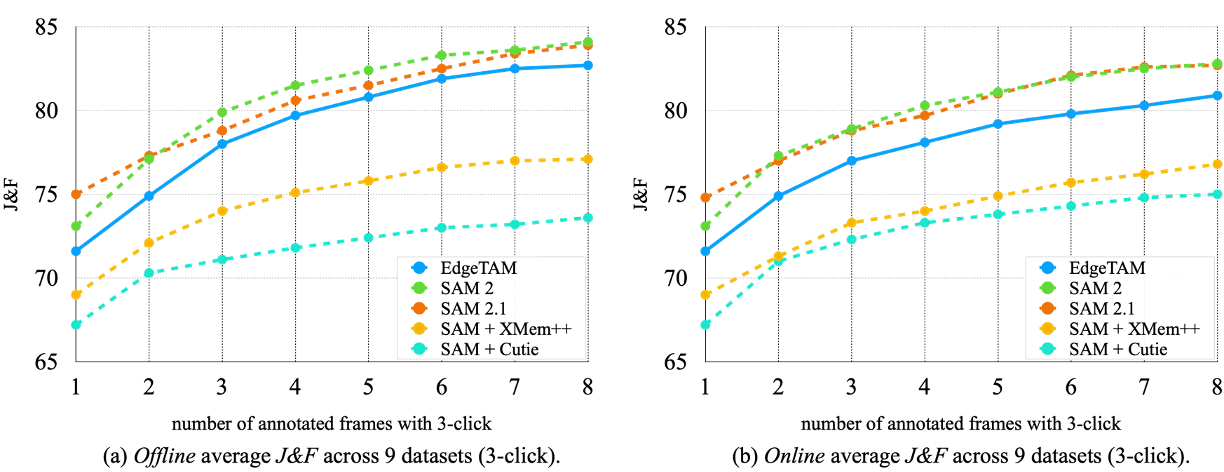}
    \caption{\textbf{Zero-shot PVS accuracy across 9 datasets in offline and online settings.}}
    \label{fig:all_pvs}
\end{figure*}

%% file: table/1_tab_img.tex
\begin{table*}[t]
\centering
\caption{\textbf{Zero-shot accuracy on the SA task across 23 datasets.} We report 1 (5) click mIoU results. FPS is measured on iPhone. Our mix does not contain the internal datasets that SAM 2 uses.}
\begin{tabular}{ll|ccc|c}
\toprule
    \f{Model}       & \f{Data}      & \f{SA-23 All}       & \f{SA-23 Image}   & \f{SA-23 Video}   & \f{FPS} \\ 
\midrule
    SAM             & SA-1B         & 58.1 (81.3)         & 60.8 (82.1)       & 54.5 (80.3)       & - \\ 
    SAM 2           & SA-1B         & 58.9 (81.7)         & 60.8 (82.1)       & 56.4 (81.2)       & 1.3 \\ 
    SAM 2           & SAM2's mix    & 61.4 (83.7)         & 63.1 (83.9)       & 59.1 (83.3)       & 1.3 \\
    SAM 2.1         & SAM2's mix    & \f{61.9 (83.5)}     & \f{63.3 (83.8)}   & \f{60.1 (83.2)}   & 1.3 \\
\midrule
    \f{EdgeTAM}     & Our mix       & 55.5 (81.7)         & 56.0 (81.9)       & 54.8 (81.5)       & \f{40.4} \\ 
\bottomrule
\end{tabular}
\label{tab:img}
\end{table*}

%% file: table/2_tab_vos.tex
\begin{table*}[t]
\centering
\caption{\textbf{Performance on the VOS task.} We report the \g{} for YTVOS and \jnf{} for other datasets. The FPS on A100 is obtained with torch compile. Nota that, for SAM 2, SAM 2.1, and EdgeTAM, we evaluate all the datasets with the same model.}
\begin{tabular}{lccccc|ccc}
\toprule
\textbf{Method} & \f{\makecell{MOSE\\val}} & \f{\makecell{DAVIS 2017\\val}} & \f{\makecell{SA-V\\val}} & \f{\makecell{SA-V\\test}} & \f{\makecell{YTVOS 2019\\val}} & \f{A100} & \f{V100} & \f{iPhone} \\ 
\midrule
STCN \cite{stcn}                & 52.5      & 85.4       & 61.0      & 62.5      & 82.7      & 62.8    & 13.2    & - \\
SwinB-AOT \cite{aot}            & 59.4      & 85.4       & 51.1      & 50.3      & 84.5      & -       & -       & - \\
SwinB-DeAOT \cite{deaot}        & 59.9      & 86.2       & 61.4      & 61.8      & 86.1      & -       & -       & - \\
RDE \cite{rde}                  & 46.8      & 84.2       & 51.8      & 53.9      & 81.9      & 88.8    & 24.4    & - \\
XMem \cite{xmem}                & 59.6      & 86.0       & 60.1      & 62.3      & 85.6      & 61.2    & 22.6    & - \\
SimVOS-B \cite{simvos}          & -         & 88.0       & 44.2      & 44.1      & 84.2      & -       & 3.3     & - \\
JointFormer \cite{jointformer}  & -         & 90.1       & -         & -         & 87.4      & -       & 3.0     & - \\
ISVOS \cite{isvos}              & -         & 88.2       & -         & -         & 86.3      & -       & 5.8     & - \\
DEVA  \cite{deva}               & 66.0      & 87.0       & 55.4      & 56.2      & 85.4      & 65.2    & 25.3    & - \\
Cutie-base \cite{cuite}         & 69.9      & 87.9       & 60.7      & 62.7      & 87.0      & 65.0    & 36.4    & - \\
Cutie-base+ \cite{cuite}        & 71.7      & 88.1       & 61.3      & 62.8      & 87.5      & 57.2    & 17.9    & - \\
SAM 2-B+ \cite{sam2}            & 75.8      & \f{90.9}   & 73.6      & 74.1      & 88.4      & 64.8    & -       & 0.7 \\
SAM 2.1-B+ \cite{sam2}          & \f{76.6}  & 90.2       & \f{76.8}  & \f{77.0}  & \f{88.6}  & 64.1    & -       & 0.7 \\
\midrule
\f{EdgeTAM}                     & 70.0      & 87.7       & 72.3      & 71.7      & 86.2      & \f{150.9}& -      & \f{15.7} \\
\bottomrule
\end{tabular}
\label{tab:vos}
\end{table*}

%% file: table/3_tab_ablation.tex
\begin{table*}[t]
\centering
\caption{\textbf{Ablation Studies.}}
\begin{subtable}[t]{0.5\textwidth}
\centering
\caption{Effectiveness of each proposed component.}
\begin{tabular}{cc|ccc}
\toprule
\f{Memory Efficiency} & \f{Distill} & \f{\makecell{SA-V\\val}} & \f{\makecell{SA-V\\test}} & \f{FPS} \\ 
\midrule
 -                               &                           & 63.5         & 62.1          & 2.5 \\
 Average Pooling                 &                           & 61.8         & 59.8          & \f{15.7} \\
 2D Perceiver                    &                           & 64.4         & 62.5          & \f{15.7} \\
 \cellcolor{gray!25}2D Perceiver & \cellcolor{gray!25}\cmark & \f{65.7}     & \f{65.8}      & \f{15.7} \\
\bottomrule
\end{tabular}
\end{subtable}
\quad
\begin{subtable}[t]{0.45\textwidth}
\centering
\caption{Latents allocation for 2D Perceiver.}
\begin{tabular}{cc|ccc}
\toprule
\f{\makecell{Global\\Latents}} & \f{\makecell{2D\\Latents}} & \f{\makecell{SA-V\\val}} & \f{\makecell{SA-V\\test}} & \f{FPS} \\ 
\midrule
 0                      & 0                         & 63.5         & 62.1          & 2.5 \\
 256                    & 0                         & 62.0         & 60.6          & \f{15.7} \\
 0                      & 256                       & 63.1         & 62.4          & \f{15.7} \\
 \cellcolor{gray!25}256 & \cellcolor{gray!25}256    & \f{64.4}     & \f{62.5}      & \f{15.7} \\
\bottomrule
\end{tabular}
\end{subtable}
\hfill
\begin{subtable}[t]{0.5\textwidth}
\centering
\caption{EdgeTAM with different backbones and \# of memory attention blocks.}
\begin{tabular}{cc|ccc}
\toprule
\f{\makecell{Image\\Encoder}}   & \f{\makecell{Mem. Attn.\\Blocks}} & \f{\makecell{SA-V\\val}} & \f{\makecell{SA-V\\test}} & \f{FPS} \\ 
\midrule
 ViT-Tiny                       & 1                     & 65.1        & 64.1          & 8.5 \\
 ViT-Tiny                       & 2                     & \f{67.9}    & \f{66.0}      & 7.4 \\
 RepViT-M1                      & 1                     & 64.3        & 61.6          & \f{22.2} \\
 \cellcolor{gray!25}RepViT-M1   & \cellcolor{gray!25}2  & 65.7        & 65.8          & 15.7 \\
 RepViT-M1                      & 4                     & 65.0        & 65.6          & 10.0 \\
\bottomrule
\end{tabular}
\end{subtable}
\quad
\begin{subtable}[t]{0.4\textwidth}
\centering
\caption{Ablation on using self attention in 2D Perceiver.}
\begin{tabular}{c|ccc}
\toprule
\f{\makecell{Self-Attn in\\Perceiver}} & \f{\makecell{SA-V\\val}} & \f{\makecell{SA-V\\test}} & \f{FPS} \\ 
\midrule
 No                      & 62.6        & \f{62.7}       & 15.7 \\
 \cellcolor{gray!25}Yes  & \f{64.4}    & 62.5           & 15.7 \\ 
\bottomrule
\end{tabular}
\end{subtable}
\hfill

\label{tab:ablation}
\end{table*}

%% file: figure/5_fig_vis.tex
\begin{figure*}[t]
    \centering
    \includegraphics[width=1.0\textwidth]{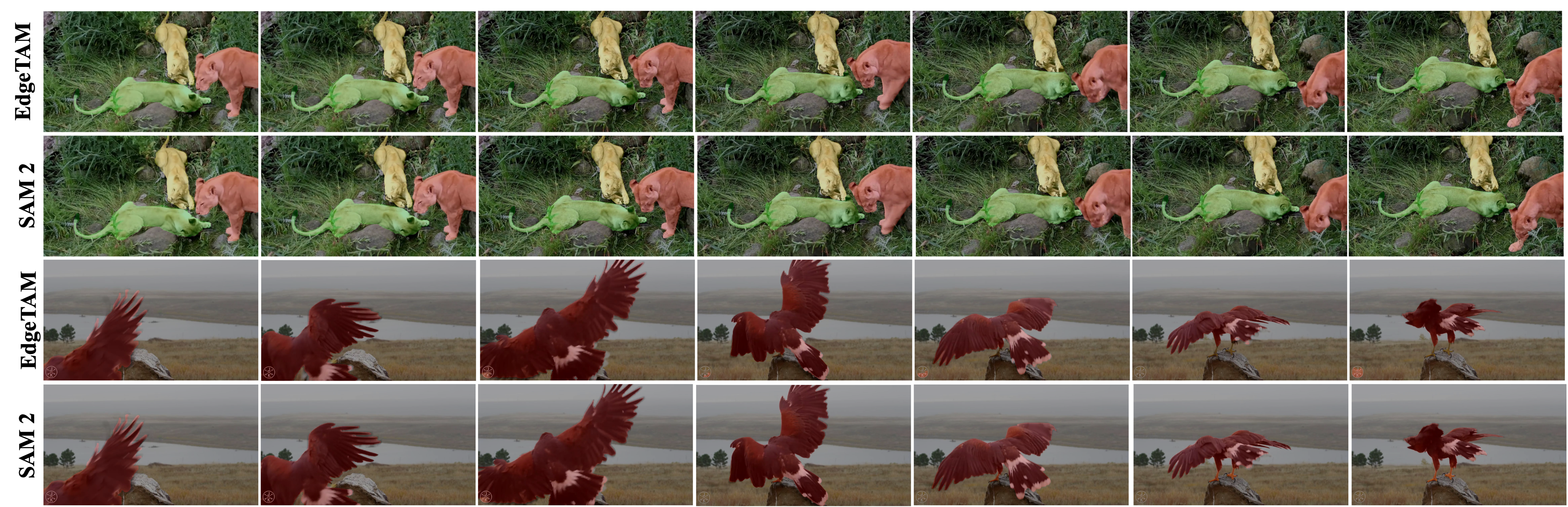}
    \caption{\textbf{Qualitative results of EdgeTAM compared with SAM 2.} In the upper example, we show tracking multiple instances from the same class, which also stay closely to each other. Our EdgeTAM delivers similar mask quality as SAM 2. In the lower example, we demonstrate a fast moving object with large distortion. While in general, EgdeTAM yields results that the boundary well, it outputs different granularities as SAM 2, not tracking the bird feet.}
    \label{fig:vis}
\end{figure*}

%% file: table/suppl/2_tab_prompt_vos.tex
\begin{table*}[t]
\centering
\caption{\textbf{Zero-shot accuracy across 17 video datasets under semi-supervised VOS evaluation using different prompts.} For all prompt types, the annotation is only provided on the first frame. \crs{}: When the ground-truth mask is available, SAM is not used for XMem++ and Cuite.}
\begin{tabular}{lccccc}
\toprule
\textbf{Method}               & \textbf{1-click} & \textbf{3-click} & \textbf{5-click} & \textbf{bounding box} & \textbf{ground-truth mask\crs} \\
\midrule
SAM + XMem++ \cite{xmem++}      & 56.9      & 68.4       & 70.6      & 67.6      & 72.7 \\
SAM + Cutie \cite{cuite}        & 56.7      & 70.1       & 72.2      & 69.4      & 74.1 \\
SAM 2 \cite{sam2}               & 64.3      & 73.2       & 75.4      & 72.9      & 77.6 \\
SAM 2.1 \cite{sam2}             & \f{64.7}  & \f{75.3}   & \f{77.6}  & \f{74.4}  & \f{79.3} \\
\midrule
\f{EdgeTAM}                     & 54.4      & 72.7      & 75.5       & 71.3      & 77.0 \\
\bottomrule
\end{tabular}
\label{tab:prompt_vos}
\end{table*}

%% file: sec/5_conclusion.tex
\section{Conclusion}
\label{sec:conclusion}
In this paper, we identify that the latency bottleneck of SAM 2 lies in the memory attention module and propose EdgeTAM to reduce the heavy overhead of cross attention with minimal performance degradation. Specifically, we propose 2D Spatial Perceiver to encode the densely stored frame-level memories into much smaller token sets while preserving their 2D spatial structure, which is essential for dense prediction tasks. As a plug-in module, 2D Spatial Perceiver can be applied to any SAM 2 variants. Besides, we also extend the knowledge distillation pipeline used in SAM for image segmentation to the video domain, which further improves the performance of EdgeTAM without inference-time cost. Our experiments show EdgeTAM nicely preserves the capability of SAM 2 across PVS, VOS, and SA tasks. More importantly, it runs 22$\times$ faster than SAM 2 and achieves 16 FPS on iPhone 15 Pro Max.

%% file: sec/X_suppl.tex

\setcounter{section}{0}
\renewcommand{\thesection}{\Alph{section}}

\section{Video Object Segmentation (VOS)}
In our main submission, we follow the standard semi-supervised video object segmentation protocol, where the ground-truth masks on the first frame are available during inference. In Tab.~\ref{tab:prompt_vos}, we follow SAM 2 \cite{sam2} and instead of making the masks on the first frame available, we prompt the object of interest with clicks or boxes on the first frame. Given that XMem++ and Cutie do not support these prompts, we convert the prompt to masks with SAM \cite{sam}. We evaluate on 17 zero-shot datasets including EndoVis 2018 \cite{endovis}, ESD \cite{esd}, LVOSv2 \cite{lvos}, LV-VIS \cite{lvvis}, UVO \cite{uvo}, VOST \cite{vost}, PUMaVOS \cite{xmem++}, Virtual KITTI 2 \cite{kitti}, VIPSeg \cite{vipseg}, Wildfires \cite{wildfires}, VISOR \cite{visor}, FBMS \cite{fbms}, Ego-Exo4D \cite{ego}, Cityscapes \cite{cityscapes}, Lindenthal Camera \cite{lindenthal}, HT1080WT Cells \cite{cell}, and Drosophila Heart \cite{drosophila}. 


In this evaluation suite, except for the 1-click setting, EdgeTAM surpasses the strong baselines, SAM + XMem++ and SAM + Cutie, by 2 to 5 percent. Compared to SAM 2 and SAM 2.1, EdgeTAM still preserves comparable performance especially with more accurate prompts, such as 5-click and ground-truth mask.



\input{table/suppl/4_tab_impl}

\section{Implementation Details}
We generally follow the original SAM 2 training hyper-parameters for image segmentation pre-training \cite{sam} and video segmentation training \cite{sam2}. Here, we highlight only the differences, and the full training details are shown in Tab.~\ref{tab:hyperparams}. First, we do not apply drop path or layer-wise decay in the image encoder. Second, our image pre-training stage adopts a 128 batch size and a total of 175K training steps. In the video training stage, we reduce the maximum number of masks per image from 64 to 32. More importantly, we do not train on the SAM 2 Internal dataset so the total training steps are reduced from 300K to 130K. Finally, our training involves distillation losses in both stages.

\section{Speed Benchmark}
In Tab.~\ref{tab:vos}, we provide the throughput FPS on both server GPUs (NVIDIA A100 and V100) and mobile NPU (iPhone 15 Pro Max). The V100 benchmarks are collected from each individual paper and we benchmark with the other two hardware by ourselves. In particular, to optimize the throughput, on A100, we torch compile all the models. For mobile NPU, we convert the model to CoreML format with coremltools \cite{coremltools} and benchmark with the performance report tool of XCode with iOS 18.1 on an iPhone 15 Pro Max. Note that, the speed-up ratios of EdgeTAM \emph{v.s.} SAM 2 are less pronounced on A100 than on iPhone. To understand the root cause, we monitor the streaming multiprocessor (SM) utilization of both models on A100 and find that even with torch compile, the SM usage of EdgeTAM is less than 50\% and the inference is bottlenecked on CPU and IO. We think it is because high-end server GPUs, such as A100, have an enormous amount of parallel executable units (EU) and given the tiny size of EdgeTAM, it cannot occupy all the EUs at the same time. However, the design objective of EdgeTAM is edge devices, such as mobile phones, where we see $22\times$ speed-up compared with SAM 2.


%% file: table/suppl/4_tab_impl.tex
\begin{table*}[t]
\centering
\caption{\f{Hyperparameters and details of EdgeTAM image segmentation pre-training and video segmentation training.}}

\begin{subtable}[t]{0.4\textwidth}
\caption{Image segmentation pre-training.}
\begin{tabular}{ll}
\toprule
\textbf{Config}         & \textbf{Value} \\
\midrule
data                    & SA-1B \\
steps                   & $\sim$175K \\
resolution              & 1024 \\
precision               & bfloat16 \\
optimizer               & AdamW \\
optimizer momentum      & $\beta_1, \beta_2=0.9, 0.999$ \\
gradient clipping       & type: $\ell_2$, max: 0.1 \\
weight decay            & 0.1 \\
learning rate (lr)      & $4e^{-4}$ \\
lr schedule             & reciprocal sqrt \\
                        & timescale=1000 \\
warmup                  & linear, 1K iters \\
cooldown                & linear, 5K iters \\
augmentation            & hflip \\
batch size              & 128 \\
mask losses (weight)    & focal (20), dice (1) \\
IoU loss (weight)       & $\ell_1$ (1) \\
distill loss (weight)   & MSE (1) \\
max. masks per img.     & 64 \\
\# correction points    & 7 \\
\bottomrule
\end{tabular}
\end{subtable}
\begin{subtable}[t]{0.45\textwidth}
\caption{Video segmentation training.}
\begin{tabular}{ll}
\toprule
\textbf{Config}         & \textbf{Value} \\
\midrule
data                    & SA-1B, SA-V, DAVIS, MOSE, YTVOS \\
steps                   & $\sim$130K \\
resolution              & 1024 \\
precision               & bfloat16 \\
optimizer               & AdamW \\
optimizer momentum      & $\beta_1, \beta_2=0.9, 0.999$ \\
gradient clipping       & type: $\ell_2$, max: 0.1 \\
weight decay            & 0.1 \\
learning rate (lr)      & backbone: $6e^{-5}$, other: $3e^{-4}$ \\
lr schedule             & cosine \\
warmup                  & linear, 15K iters \\
img. augmentation       & hflip \\
vid. augmentation       & hflip, \\
                        & affine (deg: 25, shear: 20), \\
                        & colorjitter (0.1), \\
                        & grayscale (0.05), \\
                        & per frame colorjitter (0.1) \\
batch size              & 256 \\
mask losses (weight)    & focal (20), dice (1) \\
IoU loss (weight)       & $\ell_1$ (1) \\
occlusion loss (weight) & cross-entropy (1) \\
distill loss (weight)   & MSE (1) for both $\mathcal{L}_\text{img}$ and $\mathcal{L}_\text{mem}$ \\
max. masks per frame    & image: 32, video: 3 \\
\# correction points    & 7 \\
\bottomrule
\end{tabular}
\end{subtable}

\label{tab:hyperparams}
\end{table*}